\documentclass{article}

\PassOptionsToPackage{numbers,compress}{natbib}
\usepackage[utf8]{inputenc} % allow utf-8 input
\usepackage[T1]{fontenc}    % use 8-bit T1 fonts
\usepackage[colorlinks=true, citecolor=black, linkcolor=black, urlcolor=blue]{hyperref}
\usepackage{url}            % simple URL typesetting
\usepackage{booktabs}       % professional-quality tables
\usepackage{amsfonts}       % blackboard math symbols
\usepackage{nicefrac}       % compact symbols for 1/2, etc.
\usepackage{microtype}      % microtypography
\usepackage{xcolor}         % colors
\usepackage{amsmath}
\usepackage{algorithm}
\usepackage{graphicx}
\usepackage{subcaption}
\usepackage{changepage}
\usepackage{algpseudocode}
\usepackage[preprint]{neurips_2025}
\usepackage{titlesec}
\titlespacing{\section}
  {0pt}{6pt plus 1pt minus 1pt}{3pt plus 1pt minus 1pt}

\titlespacing{\subsection}
  {0pt}{4pt plus 1pt minus 1pt}{2pt plus 1pt minus 1pt}

\makeatletter
\renewcommand{\@noticestring}{Accepted to AI for Science workshop (NeurIPS \@neuripsyear)}
\makeatother
\usepackage[utf8]{inputenc}
\usepackage[UKenglish]{babel}
\usepackage{csquotes}
\begin{document}
%\usepackage[backend=biber,style=vancouver,sorting=none]{biblatex}
% \addbibresource{references.bib}

\title{A study of EHVI vs fixed scalarization for molecule design}
% The \author macro works with any number of authors. There are two commands.
% used to separate the names and addresses of multiple authors: \And and \AND.
%
% Using \And between authors leaves it to LaTeX to determine where to break the
% lines. Using \AND forces a line break at that point. So, if LaTeX puts 3 of 4
% authors names on the first line, and the last on the second line, try using
% \AND instead of \And before the third author name.

\author{
  Anabel Yong\\
  Department of Computing \\
  National University of Singapore \\
  \texttt{t0936716@u.nus.edu} \\
  \And
  Austin Tripp\thanks{Work done while at University of Cambridge}\\
  Valence Labs \\
  % Senior Machine Learning Research Scientist\\
  % \\
  \texttt{austin.james.tripp@gmail.com} \\
  \And
  Layla Hosseini-Gerami \\
  Chief Data Science Officer \\
  IgnotaLabs.AI\\
  \texttt{layla.gerami@ignotalabs.ai} \\
  \And
  Brooks Paige \\
  AI Centre, Dept. of Computer Science\\
  University College London\\
  \texttt{b.paige@ucl.ac.uk} \\
}

\maketitle

\begin{abstract}
Multi-objective Bayesian optimization (MOBO) provides a principled framework for navigating trade-offs in molecular design. However, its empirical advantages over scalarized alternatives remain underexplored. We benchmark a simple Pareto-based MOBO strategy—Expected Hypervolume Improvement (EHVI)-against a simple fixed-weight scalarized baseline using Expected Improvement (EI), under a tightly controlled setup with identical Gaussian Process surrogates and molecular representations. Across three molecular optimization tasks, EHVI consistently outperforms scalarized EI in terms of Pareto front coverage, convergence speed, and chemical diversity. While scalarization encompasses flexible variants - including random or adaptive schemes—our results show that even strong deterministic instantiations can underperform in low-data regimes. These findings offer concrete evidence for the practical advantages of Pareto-aware acquisition in de novo molecular optimization, especially when evaluation budgets are limited and trade-offs are nontrivial.
\end{abstract}

\section{Introduction}
The discovery of therapeutic molecules is fundamentally a multi-objective optimization problem: a viable candidate must simultaneously satisfy competing criteria such as potency, safety, and pharmacokinetic properties \cite{nicolaou2013multi,ekins2010evolving,fromer2023computer}. Scalarization---where multiple objectives are collapsed into a single score using weighted combinations \cite{murata1996multiscalarization,helfrich2024scalarization}---remains a widely used strategy due to its compatibility with single-objective optimization pipelines. However, scalarization requires 
\textit{a priori} knowledge of objective weightings, which are often uncertain, context-dependent, or poorly defined in real-world drug design. Furthermore, any fixed weighting yields only a single point on the Pareto front, necessitating repeated and often redundant optimization runs to recover a diverse set of trade-off solutions. 

These limitations motivate Pareto-based multi-objective Bayesian optimization (MOBO) methods that preserve vector-valued structure of the problem and directly seek non-dominated solutions across all objectives \cite{fromercoleypareto2024,zhu2023sample,ahmadbelakariamoo2024}. Rather than collapsing objectives, MOBO aims to efficiently approximate the Pareto front by guiding evaluations toward regions that improve coverage. This approach has been shown to recover more chemically diverse and balanced solutions using fewer queries---particularly valuable in low-data, expensive-to-evaluate regimes. 
Beyond molecular design, MOBO has also demonstrated success across domains including materials science \cite{gopakumar2018multi,kumar2022multiobjective}, protein engineering \cite{park2022propertydag}, and robotics \cite{kim2021mo,wang2024fin}.

Despite this growing interest, few empirical studies systemically benchmark Pareto-based MOBO against specific scalarization strategies under controlled conditions. Prior work often defaults to scalarization heuristics \cite{fromer2023computer,jain2023multi} without evaluating their performance relative to the dedicated Pareto-based acquisitions. In this study, we present a controlled comparison between a fixed-weight scalarized Bayesian optimization strategy (using Expected Improvement) and a Pareto-based MOBO approach (using Expected Hypervolume Imrprovement). Both are implemented using identical Gaussian Process surrogates and molecular representations, isolating the acquisition function as the key difference. By evaluating performance across three \texttt{GUACAMOL} benchmark tasks, we highlight how even basic Pareto-based strategies can outperform scalarization in data-constrained molecular discovery scenarios---supporting MOBO as a more robust default for early-stage optimization. To ensure reproducibility, we will release the code and data upon acceptance, with links provided in the final version of the paper.

\section{Background and Related Work}
\subsection{Multi-Objective Optimization and Pareto Optimality}
Multi-objective optimization (MOO) concerns optimizing multiple competing objectives simultaneously. Formally, the goal is find $\mathbf{x}^* \in \mathcal{X}$ that maximizes a vector-valued objective
$R(\mathbf{x}) = [R_1(\mathbf{x}), \dots, R_d(\mathbf{x})]$. In general, no single $\mathbf{x}$ maximizes all objectives when they conflict. Instead, the optimal solutions form the \textit{Pareto set}, comprising all $\mathbf{x}$ for which no other $\mathbf{x}'$ improves all objectives. A point $\mathbf{x}_1$ is said to dominate $\mathbf{x}_2$ if $R_i(\mathbf{x}_1)\ge R_i(\mathbf{x}_2)$ for all $i$ and strictly greater for at least one $i$. The \textit{Pareto front} is the image of the Pareto-optimal set in objective space.

Classical techniques to approximate the Pareto front inlcude evolutionary algorithms such as NSGA-II \cite{deb2002fast} and MOEA/D \cite{zhang2007moea}, which maintain a population of candidate solutions. While effective for exploring diverse fronts, they are sample-inefficient and thus prohibitive when objective evaluations are expensive, as is often the case in scientific design problems \cite{ehrgott2005multicriteria, konakovic2020diversity}.

\subsection{Bayesian Optimization with Gaussian Processes}
Bayesian optimization is a sample-efficient framework for optimizing expensive black-box functions. It employs a surrogate model---typically a Gaussian Process ($\mathcal{GP}$)---to model the function's uncertainty and selects new evaluations using an acquisition function $\alpha(x)$ that balances exploration and exploitation \cite{brochu2010}. 

A Gaussian Process ($\mathcal{GP}$) is the most widely used surrogate in BO due to its nonparametric flexibility and closed-form posterior. A $\mathcal{GP}$ prior over functions $f\colon \mathcal{X}\to\mathbb{R}$ is specified by a mean function $m(\mathbf{x})$ (often taken as zero) and a covariance (kernel) function $k(\mathbf{x},\mathbf{x}')$ \cite{rassmussen2006}. Given observations $\mathcal{D}_n = {(\mathbf{x}_i, y_i)}_{i=1}^n$ with Gaussian noise $y_i = f(\mathbf{x}_i) +\varepsilon_i,; \varepsilon_i\sim\mathcal{N}(0,\sigma^2)$, the $\mathcal{GP}$ posterior at a candidate $\mathbf{x}$ has predictive distribution:
\begin{equation*}
    f(x) | \mathcal{D}_n \sim \mathcal{N}(\mu_n(x), \sigma^2_n(x))
\end{equation*}
where:
\begin{equation*}
    \mu_n(\mathbf{x}) = \mathbf{k}_n(\mathbf{x})^\top (K_n + \sigma^2 I)^{-1} \mathbf{y}, \quad
\sigma_n^2(\mathbf{x}) = k(\mathbf{x}, \mathbf{x}) - \mathbf{k}_n(\mathbf{x})^\top (K_n + \sigma^2 I)^{-1} \mathbf{k}_n(\mathbf{x}),
\end{equation*}
with $K_n[i,j]=k(\mathbf{x}_i,\mathbf{x}_j)$ and $\mathbf{k}_n(\mathbf{x})=[k(\mathbf{x},\mathbf{x}_i)]_{i=1}^n$ \cite{rassmussen2006}. These expressions quantify both the surrogate mean and its epistemic uncertainty, enabling a principled trade-off in $\alpha(\mathbf{x})$.

BO driven by $\mathcal{GP}$ surrogates has achieved remarkable sample efficiency across applications ranging from engineering design to hyperparameter tuning, often requiring orders of magnitude fewer evaluations than grid search or evolutionary algorithms \cite{brochu2010}. In molecular optimization, one must also capture chemical similarity in the kernel choice. A popular choice is the \emph{Tanimoto kernel}\cite{tanimoto_kernelsralaivola} on binary molecular fingerprints $x,x'\in [{0,1}]^d$, defined as:
\begin{equation*}
    k_{Tanimoto}(x,x') = \frac{x^\top x'}{||x||_2^2 + ||x'||^2_2 - x^\top x'}
\end{equation*}
which measures the ratio of shared substructures to the total fingerprint bits \cite{rogersecfp2010}. When used within a $\mathcal{GP}$, the Tanimoto kernel effectively models structure–property relationships in small-molecule spaces and yields strong predictive performance on tasks like binding affinity and solubility prediction \cite{gao2022pmo, trippgregg2021}.

\subsection{Scalarization and Pareto-based Optimization Strategies}
Methods for multi-objective optimization (MOO) broadly fall into scalarization-based and Pareto-based approaches. Scalarization methods reduce a vector-valued objective $f(x) = (f_1(x), \cdots, f_k(x))$ to a scalar-valued surrogate, enabling the use of well-established single-objective acquisition functions such as Expected Improvement (EI) or Upper Confidence Bound (UCB). A classical formulation is the weighted sum $f_{\text{ws}}(x) = \sum_{i=1}^k w_i f_i(x)$ where the weights $w_i$ reflect trade-off preferences. While widely used \cite{ehrgott2005multicriteria, knowles2006parego}, this approach is only guaranteed to recover Pareto-optimal points when the front is convex. Covering non-convex regions typically requires multiple optimization runs with diverse weight configurations, making the approach computationally expensive and potentially redundant. 

To overcome these limitations, alternative scalarization techniques such as the Tchebycheff scalarization have been developed, with $f_{Tchebycheff}(x) = \max_{1 \leq i \leq m} \{\lambda_i(f_i(x) = z^*_i)\}$ where $\lambda_i$ are scaling weights and $z_i^*$ is a reference point (typically the ideal vector of component-wise maxima). This scalarization emphasizes the worst-performing objective, yielding stronger guarantees for recovering solutions on convex fronts \cite{tchebysheff1976, smoothsettchebysheff2025}. Recent work has shown that only a small set of well-chosen Tchebycheff scalarizations can approximate the entire Pareto front with high fidelity and sample efficiency \cite{smoothsettchebysheff2025}. 

A particularly impactful advancement comes from the \emph{hypervolume scalarization framework} introduced by %Golovin and Zhang (2020)
\citet{golovin2020randomhypervolumescalarizationsprovable}, which establishes a formal connection between scalarized objectives and the hypervolume indicator---a gold-standard, Pareto-compliant metric~\cite{zitzler1998multiobjective}. Specifically, they show that for a suitable distribution \( \mathcal{D}_\lambda \) over weight vectors and corresponding scalarization functions \( s_\lambda \), the hypervolume of a set \( Y \subset \mathbb{R}^d \) with respect to a reference point \( z \in \mathbb{R}^d \) can be expressed as $HV_z(Y) = c_k \, \mathbb{E}_{\lambda \sim \mathcal{D}_\lambda} \left[ \max_{y \in Y} s_\lambda(y - z) \right]$.

This theoretical result provides a principled mechanism for converting hypervolume maximization into a sequence of scalar optimization subproblems. In practice, it enables provable convergence to the Pareto front using standard Bayesian optimization techniques, such as Thompson Sampling or UCB, by simply sampling a new scalarization $s_\lambda$ at each iteration. %Golovin and Zhang (2020) 
\citet{golovin2020randomhypervolumescalarizationsprovable} derive cumulative hypervolume regret bounds of order $\tilde{O}(T^{-1/2})$, establishing the method as both theoretically grounded and sample-efficient for multi-objective black-box optimization.

In contrast to scalarization, Pareto-based acquisition functions preserve the vectorized nature of objectives and directly aim to improve coverage of the Pareto front. Each objective is modelled independently, often using Gaussian Processes and acquisitions are scored based on their \textit{expected contribution} to the Pareto frontier. 

The most widely used Pareto-based acquisition is the Expected Hypervolume Improvement (EHVI) \cite{emmerich2011hypervolume, daulton2020differentiable}, which quantifies the increase in dominated volume achieved by adding a new sample. Formally, for the current approximation set $\mathcal{P}_t$ and reference point $z_i$, the EHVI  at candidate point $x$ is given by:
\begin{equation*}
    \text{EHVI}(x) = \mathbb{E}[HV_z(\mathcal{P}_t \cup \{f(x)\}) -HV_z(\mathcal{P_t})]
\end{equation*}

Unlike scalarization, which converts MOO into a single-objective landscape, EHVI retains the multi-dimensional structure of the problem, promoting exploration in underrepresented Pareto regions. Refinements, such as Predictive Entropy Search for Multi-objective Optimization (PESMO) \cite{hernandez-lobato-pesmo}, take an information-theoretic view, maximizing expected information gain about the Pareto set. Methods such as DGEMO \cite{konakovic2020diversity} go further by jointly optimizing for hypervolume improvement and diversity in both design and objective space. By modeling the Pareto front as a piecewise manifold and partitioning it into local diversity regions, DGEMO selects a diverse batch of samples to improve coverage efficiency - an especially useful property in low-data regimes. 
\section{Experimental Setup}
Our experiments are designed to assess the practical performance of \emph{Pareto-based acquisition strategies} for multi-objective Bayesian optimization (MOBO) in molecular design tasks. Rather than contrasting MOBO with scalarization broadly---which encompasses a diverse array of formulations including random scalarizations~\cite{golovin2020randomhypervolumescalarizationsprovable, paria2019flexibleframeworkmultiobjectivebayesian}---we focus on a controlled comparison between two representative acquisition strategies: Expected Hypervolume Improvement (EHVI), which explicitly targets Pareto front expansion, and a fixed-weight Expected Improvement (EI), a baseline scalarized acquisition function. We aim to answer the following questions: 
\begin{enumerate}
    \item \textbf{Optimality: }Does EHVI discover solutions closer to approximate  Pareto front than scalarized EI?
    \item \textbf{Diversity: }Do molecules selected by EHVI exhibit greater structural diversity?
    \item \textbf{Trade-offs: }How do EHVI and EI differ in balancing optimality and diversity under fixed BO evaluation budgets?
\end{enumerate}
We obtain positive empirical results across the 3 MPOs. 

\subsection{Benchmark Tasks}
We evaluate both methods on three multi-property optimization tasks from \texttt{GUACAMOL} \cite{brown2019guacamol}: Amlodipine MPO, Fexofenadine MPO, and Perindopril MPO. Each task involves optimizing three molecular properties jointly—target similarity, QED, and either logP, SA score, or molecular weight—thus posing realistic trade-offs encountered in drug discovery pipelines.

\subsection{Optimization Setup}

We adopt a multi-objective Bayesian optimization framework where each molecular property $f_j(m)$ is modeled independently using Gaussian Processes ($\mathcal{GP}$s). Each $\mathcal{GP}$ is equipped with the MinMax kernel, a count-aware generalization of the Tanimoto kernel suitable for Morgan fingerprints. The kernel is defined as:
\begin{equation*}
    k_{\text{MinMax}}(x, x') = \frac{\sum_i \min(x_i, x_i')}{\sum_i \max(x_i, x_i')}
\end{equation*}
This kernel measures structural similarity between molecules encoded as extended-connectivity fingerprints (ECFPs) \cite{rogersecfp2010} of radius 3, computed via RDKit using count-based features without truncation. Each molecular property is modeled as a $\mathcal{GP}$ prior $f_j \sim \mathcal{GP}(\mu_j, K_j(x_i, x_q))$. Predictions across all objectives yield independent Gaussian posteriors characterized by predictive means and variances $\vec{\mu}(m)$ and $\vec{\sigma}^2(m)$, respectively. The predictive distribution for each $\mathcal{GP}$ is Gaussian, parameterized by a mean and variance derived from exact kernel matrix computations. These $\mathcal{GP}$s are implemented in a JAX-based framework, \texttt{kernel\_only\_GP}, supporting efficient parallelization and differentiable matrix operations. All model hyperparameters, including amplitude $\alpha = 1.0$ and noise variance $s = 10^{-4}$, are held fixed across trials and methods to ensure consistent modeling assumptions.

We compare two acquisition strategies: Expected Hypervolume Improvement (EHVI), a Pareto-based method that promotes non-dominated frontier expansion, and scalarized Expected Improvement (EI) with fixed weights, which collapses the objectives into a single scalar score. EHVI is computed via Monte Carlo sampling using 1000 draws per candidate molecule. At each optimization step, a single molecule is selected from a fixed candidate pool of 10,000 compounds sampled from the \texttt{GUACAMOL} training set. The selected molecule is evaluated, added to the training archive, and the $\mathcal{GP}$ models are updated accordingly. Each run proceeds for 200 optimization rounds.

Experiments are repeated with three different random seeds per method. We report the mean and standard deviation of all evaluation metrics. Given the small number of trials, we assess the consistency and magnitude of observed differences using effect size metrics—Cohen’s $d$ \cite{cohen2013statistical} and Cliff’s Delta \cite{cliff2014ordinal}. All computations are performed on NVIDIA H100-47 GPUs.

\subsection{Evaluation Metrics and Performance Indicators}
To evaluate the performance of multi-objective optimization across both convergence quality and molecular diversity, we adopted three complementary metrics: Hypervolume Indicator (HVI) \cite{fonsecahypervolume}, the R2 indicator \cite{hansenr2} and the \#Circles metric \cite{xie2023spaceexploredmeasuringchemical}. The HVI measures the volume of the objective space dominated by the non-dominated solutions relative to a reference point, capturing both convergence to the Pareto front and diversity of trade-offs explored. A higher HVI value indicates broader and more optimal coverage of the objective space. The $R^2$ indicator quantifies the quality of the Pareto front approximation by comparing it to a fixed set of uniformly distributed reference directions, as adapted from the setup in %Jain et al(2023)'s
\citet{jain2023multi}. 
%setup. 
Specifically, it computes the augmented Tchebycheff scalarization in each direction: for each reference vector $v$, it calculates the worst-case deviation $\text{max}_i v_i \cdot |u_i - s_i|$, where $u$ is the utopian (ideal) point and $s$ is a candidate solution. It then selects the best such solution for each direction and averages over all directions. Thus, lower $R^2$ values indicate a front that is both closer and more uniformly distributed relative to the ideal front \cite{hansenr2}. Finally, the \#Circles metric quantifies structural diversity in the chemical space by counting the number of pairwise dissimilar molecules exceeding a Tanimoto distance threshold $t$. We compute it on the set of Pareto-optimal candidates from the initial and acquired molecules over 200 BO evaluations, as the metric is designed to assess the effective space explored by a \textit{representative} set of high-quality or relevant solutions rather than all generated samples \cite{xie2023spaceexploredmeasuringchemical}. Informed by axiomatic diversity principles, \#Circles offers a geometry-aware, thresholded view of chemical diversity: at higher thresholds (i.e. requiring greater dissimilarity), a larger \#Circles value indicates that the candidate set spans more structurally distinct regions of the chemical space. 

\section{Results}
This section presents a comparative evaluation of EHVI and scalarized EI across 3 multi-objective molecular optimization tasks. Across all tasks and metrics - hypervolume (Figure \ref{fig:hvi_all}), $R^2$ indicator (Figure \ref{fig:r2_all}), and chemical diversity via the \#Circles metric (Figure \ref{fig:circles_all}), EHVI demonstrates consistent advantages. Statistical tests provided  in Appendix \ref{appendix2} (Tables \ref{tab:hv_effect_sizes}, \ref{tab:r2_effectsize}) corroborate these trends: for hypervolume, EHVI achieves medium to large effect sizes (Cohen's $d = 0.576-1.093$) and favorable Cliff's delta values indicating better performance across matched trials. For the $R^2$ indicator, lower values for EHVI suggest improved approximation of the Pareto front, with strong negative effect sizes (e.g. $d = -2.56$ on Fexofenadine MPO). In terms of diversity, EHVI consistently explores more structurally distinct solutions at higher Tanimoto thresholds. These results demonstrate EHVI's robustness: it not only accelerates convergence and enhances front coverage, it also maintains better front approximation and promotes chemically diverse candidate solutions. We next present the detailed metric-wise breakdown across all tasks. 

\subsection{Hypervolume Indicator (HVI)}
\begin{figure}[H]
    \centering
    % \begin{adjustwidth}{-2.5cm}{-2.5cm}  % Extend only the figure content into the margins
        \begin{subfigure}[t]{0.33\textwidth}
            \includegraphics[width=\textwidth]{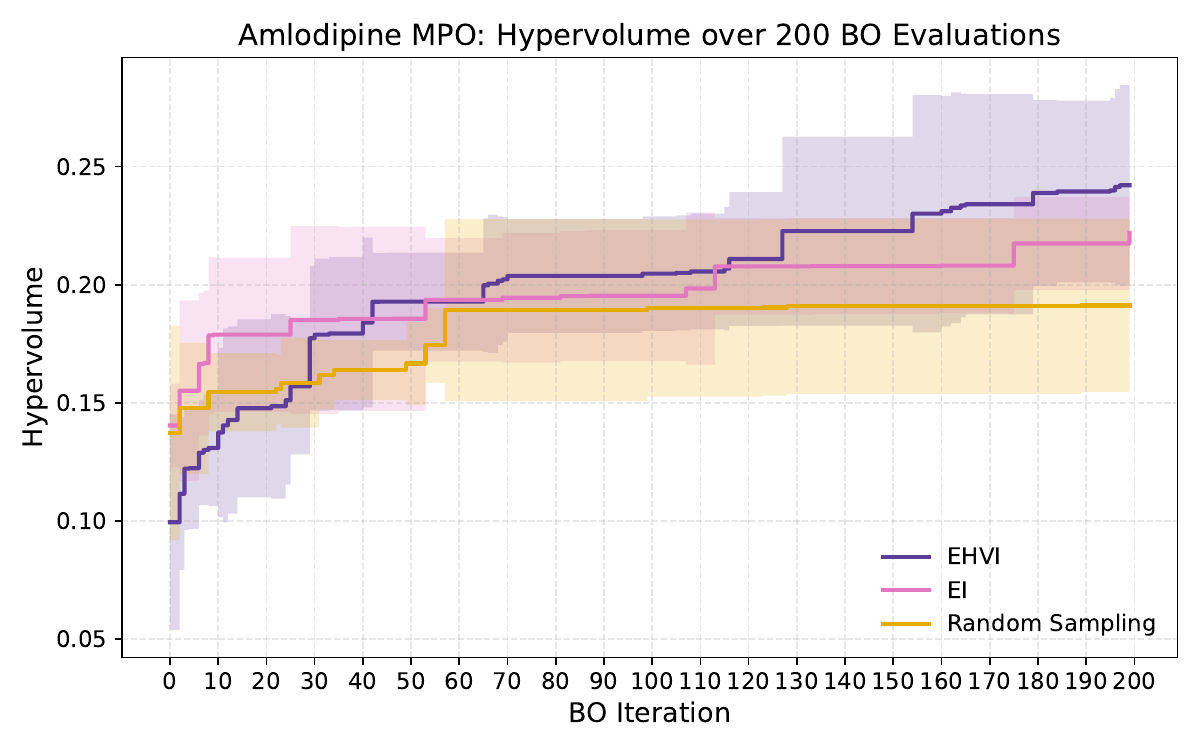}
            \caption{Amlodipine MPO}
        \end{subfigure}
        \hfill
        \begin{subfigure}[t]{0.32\textwidth}
            \includegraphics[width=\textwidth]{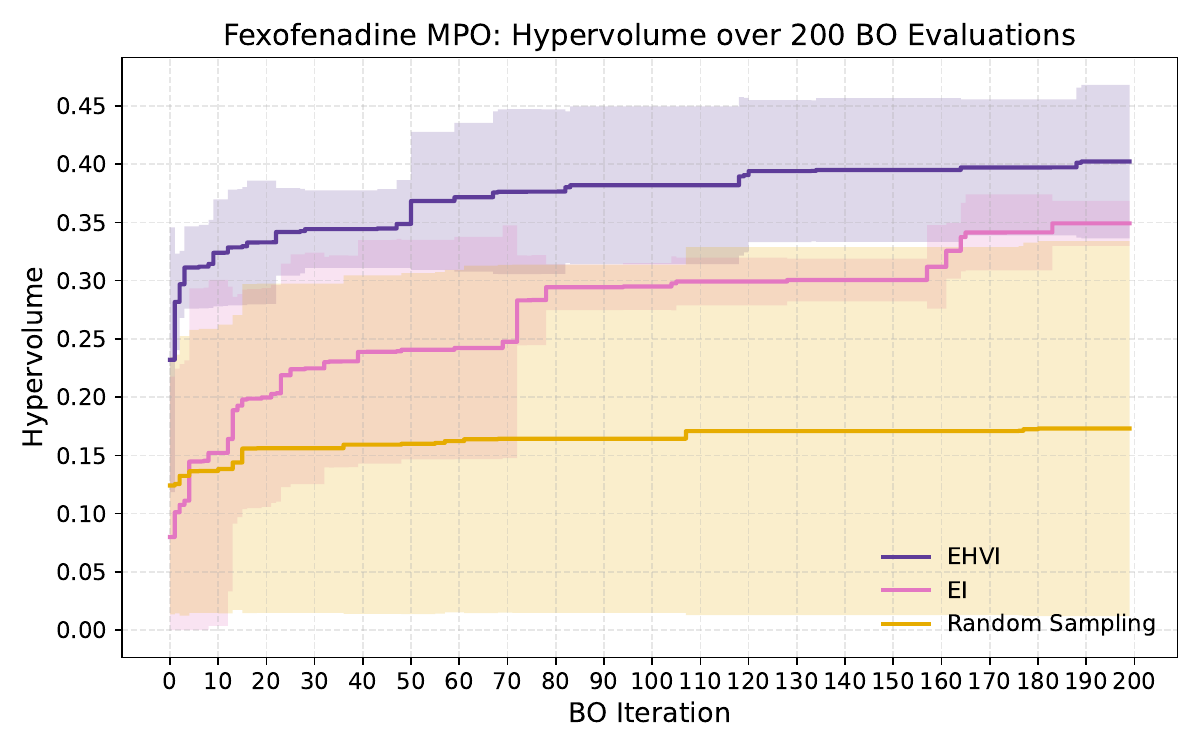}
            \caption{Fexofenadine MPO}
        \end{subfigure}
        \hfill
        \begin{subfigure}[t]{0.33\textwidth}
            \includegraphics[width=\textwidth]{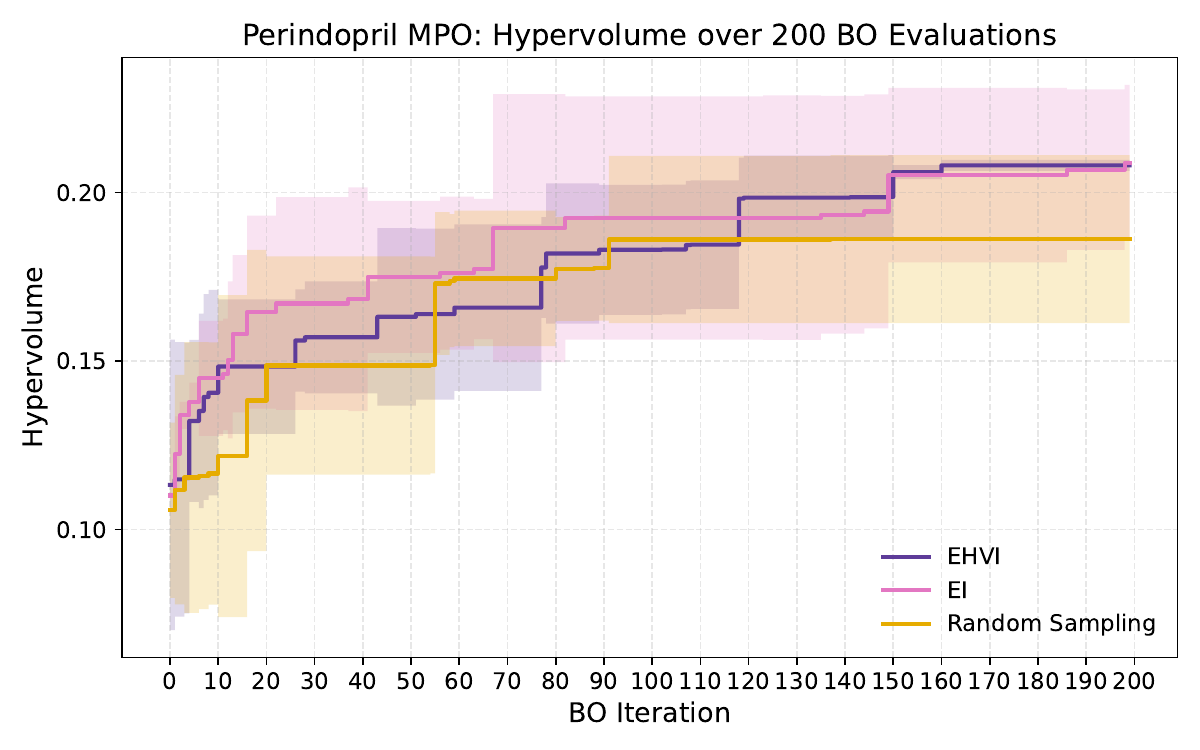}
            \caption{Perindopril MPO}
        \end{subfigure}
    % \end{adjustwidth}
    \caption{Hypervolume indicator (HVI) over 200 Bayesian optimization iterations for each MPO task. EHVI consistently achieves higher hypervolume than scalarized EI and random sampling, with faster convergence and greater final front coverage. Shaded areas represent standard deviation over 3 random seeds.}
    \label{fig:hvi_all}
\end{figure}

Across all tasks, EHVI consistently outperforms scalarized EI and random sampling in hypervolume performance (Figure \ref{fig:hvi_all}. In Amlodipine, EHVI achieves faster convergence and higher final hypervolume with lower variance. For Fexofenadine, the gap is even more pronounced—EHVI dominates throughout, especially in later stages. In Perindopril, both methods reach similar final values, but EHVI converges earlier and exhibits reduced variance, indicating greater sample efficiency and robustness. These results underscore EHVI’s consistent advantage in front expansion and reliability across diverse multi-objective settings.

\subsection{\texorpdfstring{$R^2$}{R2} Indicator}
In Figure \ref{fig:r2_all} below, EHVI exhibits clear superiority in Pareto front approximation across all tasks. In Fexofenadine, it achieves the lowest and most stable $R^2$ scores, with a wide and persistent gap over scalarized EI. Amlodipine shows modest separation, with EHVI attaining consistently lower scores after early fluctuations. In Perindopril, EHVI dominates from mid-optimization onward, yielding the most stable and lowest variance estimates. These trends indicate that EHVI not only expands the front efficiently but also produces more uniformly optimal trade-offs across objectives.
\begin{figure}[H]
    \centering
    % \begin{adjustwidth}{-2.5cm}{-2.5cm}  % Extend only the figure content into the margins
        \begin{subfigure}[t]{0.33\textwidth}
            \includegraphics[width=\textwidth]{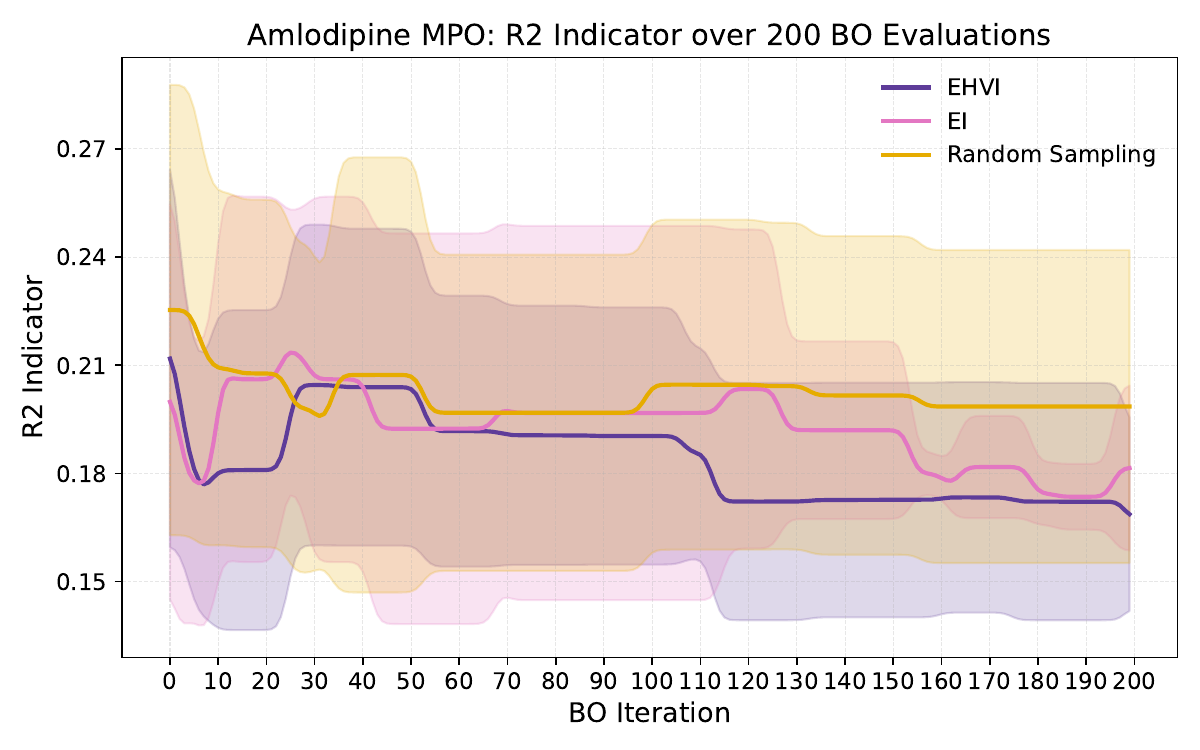}
            \caption{Amlodipine MPO}
        \end{subfigure}
        \hfill
        \begin{subfigure}[t]{0.32\textwidth}
            \includegraphics[width=\textwidth]{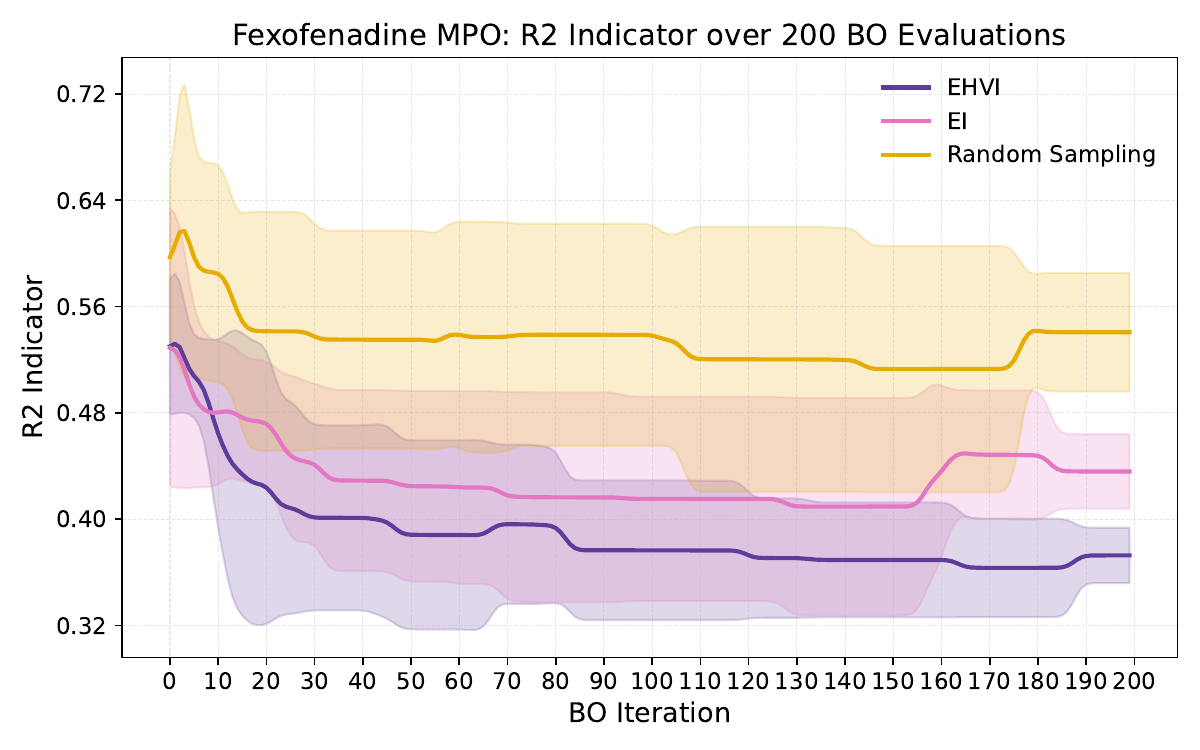}
            \caption{Fexofenadine MPO}
        \end{subfigure}
        \hfill
        \begin{subfigure}[t]{0.33\textwidth}
            \includegraphics[width=\textwidth]{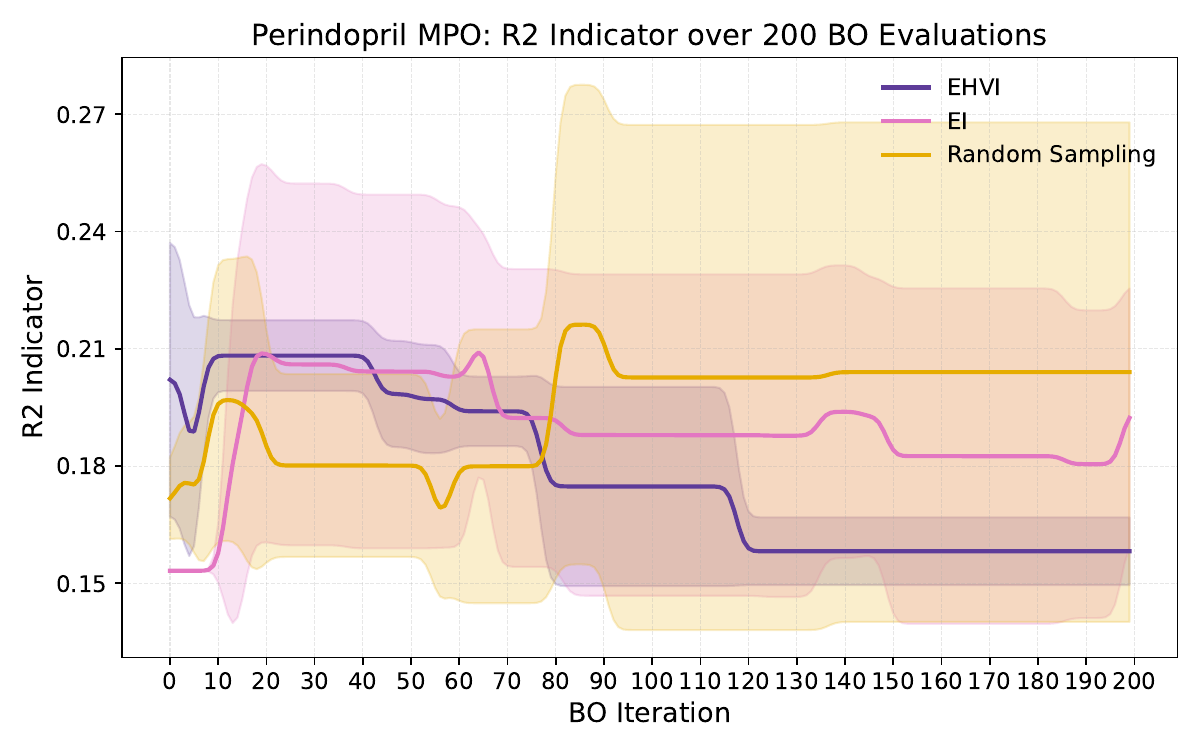}
            \caption{Perindopril MPO}
        \end{subfigure}
    % \end{adjustwidth}
    \caption{$R^2$ indicator across 200 Bayesian optimization iterations for each MPO task. Lower values reflect better approximation of the true Pareto front under varying utility directions. EHVI consistently achieves lower $R^2$ values than scalarized EI and random sampling, indicating superior convergence toward the reference front. Shaded regions show standard deviation over 3 random seeds.}
    \label{fig:r2_all}
\end{figure}

\subsection{\#Circles Metric}
\begin{figure}[H]
    \centering
    % \begin{adjustwidth}{-2.5cm}{-2.5cm}  % Extend only the figure content into the margins
        \begin{subfigure}[t]{0.33\textwidth}
            \includegraphics[width=\textwidth]{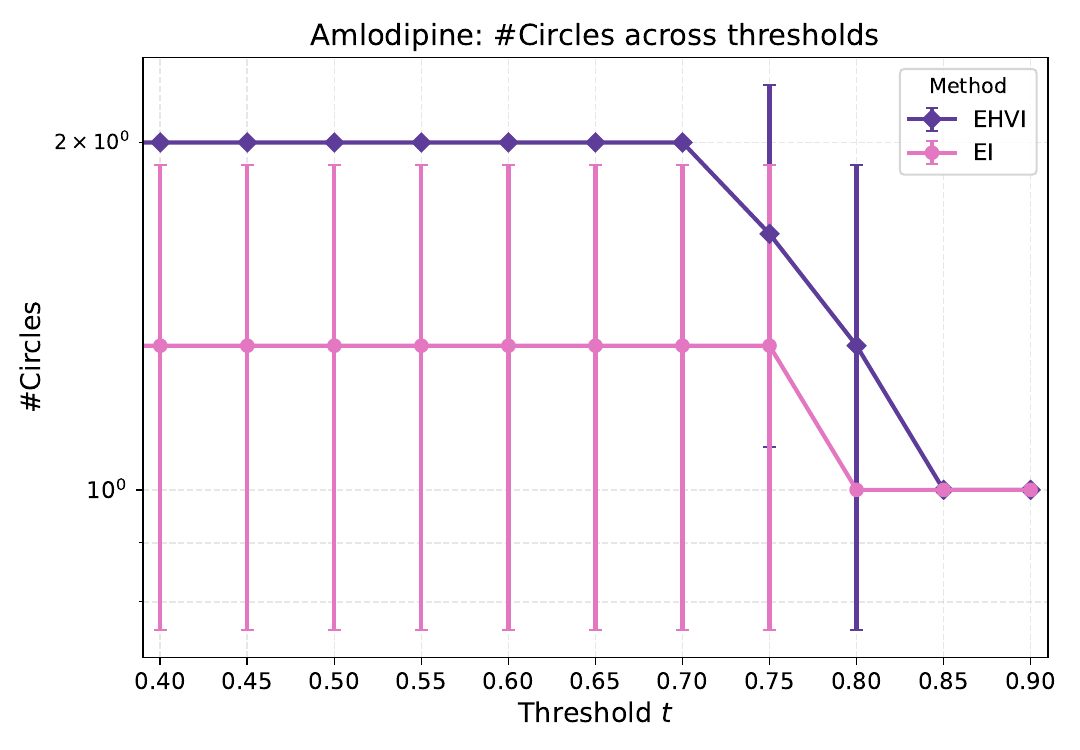}
            \caption{Amlodipine MPO}
        \end{subfigure}
        \hfill
        \begin{subfigure}[t]{0.32\textwidth}
            \includegraphics[width=\textwidth]{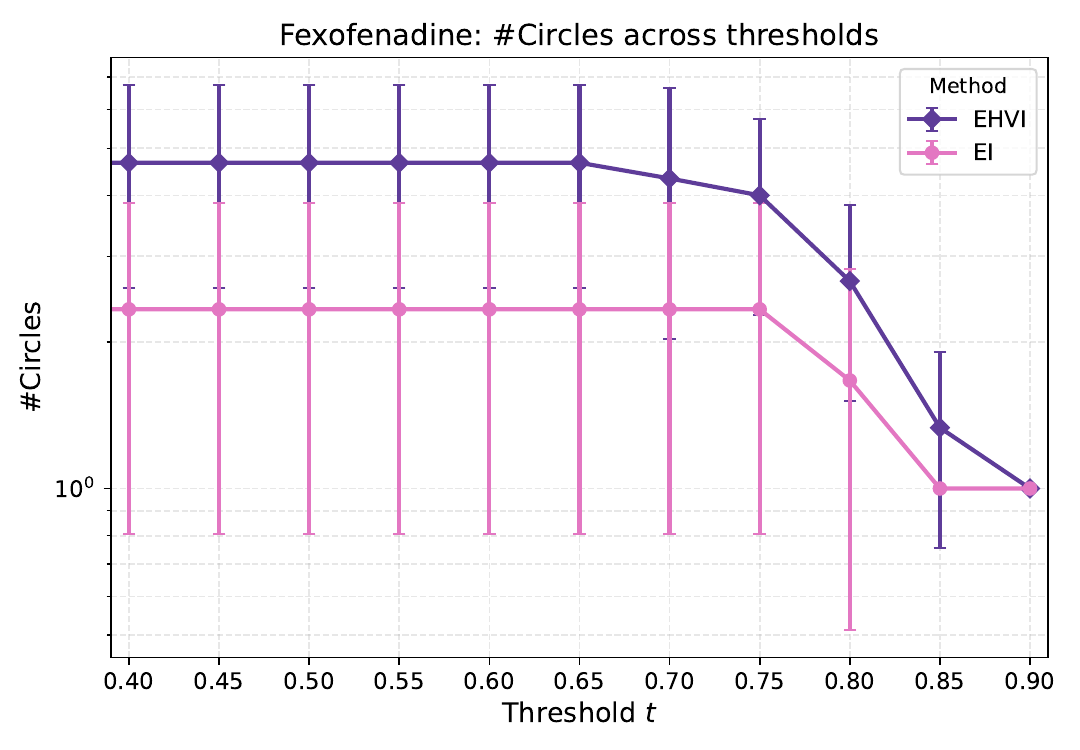}
            \caption{Fexofenadine MPO}
        \end{subfigure}
        \hfill
        \begin{subfigure}[t]{0.33\textwidth}
            \includegraphics[width=\textwidth]{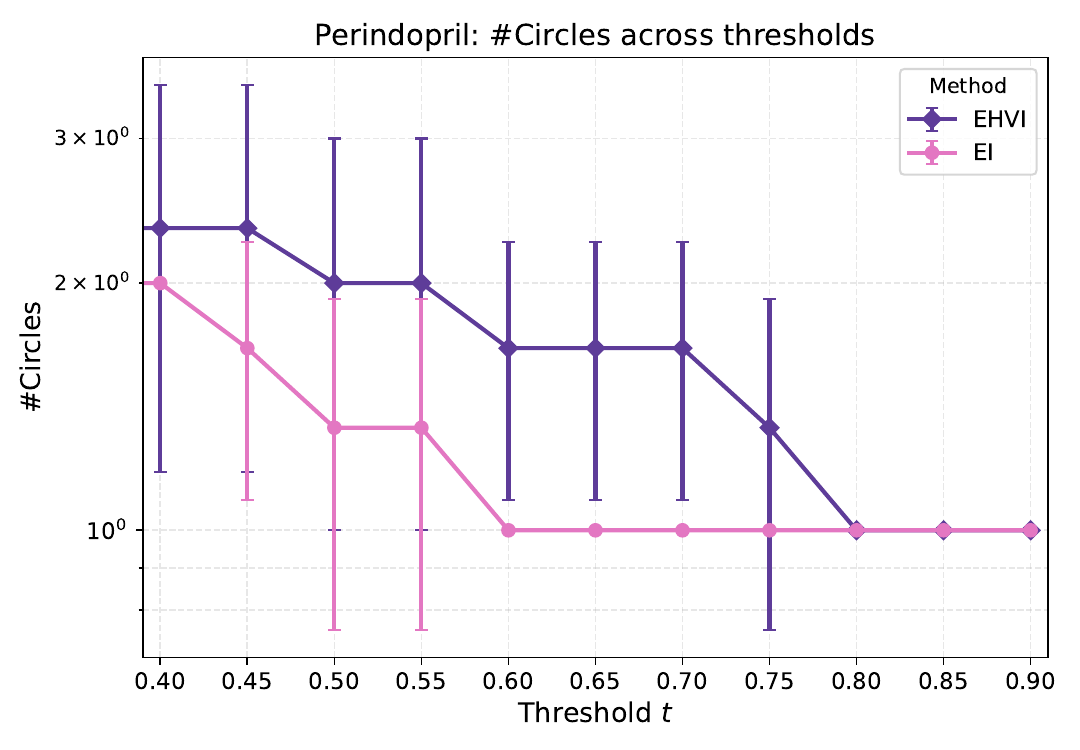}
            \caption{Perindopril MPO}
        \end{subfigure}
    % \end{adjustwidth}
    \caption{Structural diversity assessed using the \#Circles metric across increasing Tanimoto distance thresholds. Higher values indicate broader exploration of structurally distinct regions of the chemical space. EHVI consistently maintains or exceeds the diversity of scalarized EI, particularly at stricter thresholds. Error bars denote standard deviation across 3 random seeds.}
    \label{fig:circles_all}
\end{figure}

EHVI demonstrates superior or comparable chemical diversity across all MPO tasks shown in Figure \ref{fig:circles_all}. In Fexofenadine, EHVI clearly outpaces EI at thresholds $t \geq 0.60$, uncovering significantly more distinct structural motifs. Perindopril shows a similar advantage, with EHVI sustaining diversity across the full range of thresholds. In Amlodipine, both methods perform similarly at low to mid thresholds, but EHVI maintains higher diversity beyond $t=0.75$ suggesting enhanced exploration of chemically dissimilar optima. These patterns highlight EHVI’s ability to balance objective performance with structural novelty.

\section{Discussion}
This study offers a systematic empirical investigation of Pareto-based Bayesian optimization using Expected Hypervolume Improvement (EHVI), contrasting it against a widely used single-objective acquisition strategy: scalarized Expected Improvement (EI) with fixed weights. We focus on a fixed-weight baseline, commonly used in practical pipelines. This allows us to isolate the benefits of Pareto-aware acquisition in molecular optimization, without confounding from scalarization strategy variation.

Our findings demonstrate that EHVI yields more sample-efficient exploration across a range of realistic multi-property optimization (MPO) tasks. These gains are reflected across three orthogonal metrics—hypervolume coverage, front approximation accuracy ($R^2$), and structural diversity—and are statistically robust across random seeds, as shown in Appendix \ref{appendix2} (Tables \ref{tab:hv_effect_sizes}, \ref{tab:r2_effectsize}). Crucially, these improvements are not due to changes in surrogate fidelity, kernel choice, or representation capacity: both EHVI and scalarized EI operate under the same model assumptions and input features, underscoring the importance of acquisition strategy alone in driving performance.

Another promising direction involves evaluating the role of Monte Carlo (MC) sampling in EHVI’s performance. Our current setup uses 1000 MC samples per candidate evaluation, which strikes a balance between computational overhead and estimate fidelity. However, increasing this number could reduce integration variance and enhance the accuracy of hypervolume estimates, potentially accelerating convergence and improving solution quality. A dedicated ablation study could clarify whether EHVI disproportionately benefits from higher sampling rates—particularly in higher-dimensional tasks where MC estimation errors can compound \cite{lepage-mcintegration}. Similarly, future work should investigate how the precision of the Monte Carlo estimate—i.e., the amount of stochastic noise in the acquisition value—impacts optimization behavior. In particular, it remains unclear whether EHVI or scalarized EI are more sensitive to noisy acquisition signals, which could affect candidate selection and convergence speed.

We also emphasize the role of molecular representation. This study employed full-dimensional count-based ECFP vectors with a MinMax kernel—an information-rich but high-dimensional descriptor space. While this aligns with recent findings that preserving fingerprint fidelity improves surrogate model accuracy \cite{tripp2024diagnosingfixingcommonproblems}, it also introduces challenges related to model scalability and sample efficiency. Benchmarking EHVI and EI using reduced-dimensional or contrastively learned fingerprint embeddings could reveal how robust each method is to representation compression or abstraction.

Despite the promising results presented here, our surrogate models use independent Gaussian Processes without adaptive hyperparameter tuning. This choice prioritizes control and reproducibility, but may limit adaptability to evolving posterior landscapes, particularly in complex or noisy objective settings. Exploring more expressive surrogates—such as deep kernel methods, ensembling, or dynamically updated GPs—could enhance fidelity and robustness. Additionally, while EHVI and scalarized EI offer strong baselines, other acquisition strategies deserve future evaluation. For example, ParEGO \cite{knowles2006parego}, PESMO \cite{hernandez-lobato-pesmo}, and generative diversity-promoting methods like Multi-Objective GFlowNets \cite{jain2023multi} offer orthogonal strengths. Expanding evaluations to include noisy, constrained, or synthesis-feasible objectives would further establish the practical utility of Pareto-aware acquisition.

\section{Conclusion}
We conducted a focused comparison between Pareto-based Bayesian optimization (EHVI) and scalarized Expected Improvement (EI) with fixed weights across multi-objective molecular design tasks. EHVI consistently outperforms EI, achieving higher hypervolume, lower $R^2$ values, and greater structural diversity. Statistical effect size analysis further confirms these gains are meaningful despite limited runs. Unlike scalarized EI, which optimizes a fixed utility, EHVI explicitly captures trade-offs, leading to more balanced and diverse Pareto fronts. These results demonstrate the value of Pareto-aware acquisition in data-constrained settings and motivate future work on adaptive surrogates, representation learning, and diversity-aware BO strategies.

\section*{Acknowledgements}

This work was conducted as part of the first author’s Master’s thesis at University College London. We thank the thesis supervisors and committee members for their guidance and feedback.

\newpage
% \printbibliography
\bibliographystyle{plainnat}
\bibliography{references}

@article{fromer2023computer,
  title={Computer-aided multi-objective optimization in small molecule discovery},
  author={Fromer, Jenna C and Coley, Connor W},
  journal={Patterns},
  volume={4},
  number={2},
  year={2023},
  publisher={Elsevier},
  url={https://www.cell.com/patterns/fulltext/S2666-3899(23)00001-6}
}

@article{nicolaou2013multi,
  title        = {Multi-objective optimization methods in drug design},
  author       = {Nicolaou, Christos A and Brown, Nathan},
  journal      = {Drug Discovery Today: Technologies},
  volume       = {10},
  number       = {3},
  pages        = {e427--e435},
  year         = {2013},
  publisher    = {Elsevier},
  doi          = {10.1016/j.ddtec.2013.02.001}
}

@article{ekins2010evolving,
  title        = {Evolving molecules using multi-objective optimization: applying to ADME/Tox},
  author       = {Ekins, Sean and Honeycutt, J Dana and Metz, James T},
  journal      = {Drug Discovery Today},
  volume       = {15},
  number       = {11-12},
  pages        = {451--460},
  year         = {2010},
  publisher    = {Elsevier},
  doi          = {10.1016/j.drudis.2010.04.003},
}

@article{murata1996multiscalarization,
  title        = {Multi-objective genetic algorithm and its applications to flowshop scheduling},
  author       = {Murata, Tadahiko and Ishibuchi, Hisao and Tanaka, Hideo},
  journal      = {Computers \& Industrial Engineering},
  volume       = {30},
  number       = {4},
  pages        = {957--968},
  year         = {1996},
  publisher    = {Elsevier},
  doi          = {10.1016/0360-8352(96)00045-9},
  url          = {https://doi.org/10.1016/0360-8352(96)00045-9}
}

@article{helfrich2024scalarization,
  author    = {Helfrich, S. and Herzel, A. and Ruzika, S.},
  title     = {Using scalarizations for the approximation of multiobjective optimization problems: Towards a general theory},
  journal   = {Mathematical Methods of Operations Research},
  volume    = {100},
  pages     = {27--63},
  year      = {2024},
  doi       = {10.1007/s00186-023-00823-2},
  url       = {https://doi.org/10.1007/s00186-023-00823-2},
  publisher = {Springer}
}

@article{fromercoleypareto2024,
    author ="Fromer, Jenna C. and Graff, David E. and Coley, Connor W.",
    title  ="Pareto optimization to accelerate multi-objective virtual screening",
    journal  ="Digital Discovery",
    year  ="2024",
    volume  ="3",
    issue  ="3",
    pages  ="467-481",
    publisher  ="RSC",
    doi  ="10.1039/D3DD00227F",
    url  ="http://dx.doi.org/10.1039/D3DD00227F"
}

@inproceedings{ahmadbelakariamoo2024,
  title     = {Pareto Front-Diverse Batch Multi-Objective Bayesian Optimization},
  author    = {Ahmadianshalchi, Alaleh and Belakaria, Syrine and Doppa, Janardhan R.},
  booktitle = {Proceedings of the AAAI Conference on Artificial Intelligence},
  volume    = {38},
  pages     = {12017--12025},
  year      = {2024},
  url       = {https://ojs.aaai.org/index.php/AAAI/article/view/28951},
  doi       = {10.1609/aaai.v38i10.28951}
}

@article{zhu2023sample,
      title={Sample-efficient multi-objective molecular optimization with gflownets},
      author={Zhu, Yiheng and Wu, Jialu and Hu, Chaowen and Yan, Jiahuan and Hou, Tingjun and Wu, Jian and others},
      journal={Advances in Neural Information Processing Systems},
      volume={36},
      pages={79667--79684},
      year={2023}
}

@article{gopakumar2018multi,
  title        = {Multi-objective optimization for materials discovery via adaptive design},
  author       = {Gopakumar, Abhijith M. and Balachandran, Prasanna V. and Xue, Dezhen and Gubernatis, James E. and Lookman, Turab},
  journal      = {Scientific Reports},
  volume       = {8},
  number       = {1},
  pages        = {3738},
  year         = {2018},
  publisher    = {Nature Publishing Group},
  doi          = {10.1038/s41598-018-21936-3},
  url          = {https://doi.org/10.1038/s41598-018-21936-3}
}

@article{kumar2022multiobjective,
  title={Multiobjective Bayesian optimization framework for the synthesis of methanol from syngas using interpretable Gaussian process models},
  author={Kumar, Avan and Pant, Kamal K and Upadhyayula, Sreedevi and Kodamana, Hariprasad},
  journal={ACS omega},
  volume={8},
  number={1},
  pages={410--421},
  year={2022},
  publisher={ACS Publications},
  url={https://doi.org/10.1021/acsomega.2c04919},
}

@article{park2022propertydag,
  title={Propertydag: Multi-objective bayesian optimization of partially ordered, mixed-variable properties for biological sequence design},
  author={Park, Ji Won and Stanton, Samuel and Saremi, Saeed and Watkins, Andrew and Dwyer, Henri and Gligorijevic, Vladimir and Bonneau, Richard and Ra, Stephen and Cho, Kyunghyun},
  journal={arXiv preprint arXiv:2210.04096},
  year={2022},
  url={https://doi.org/10.48550/arXiv.2210.04096},
}

@inproceedings{kim2021mo,
  title={Mo-bbo: Multi-objective bilevel bayesian optimization for robot and behavior co-design},
  author={Kim, Yeonju and Pan, Zherong and Hauser, Kris},
  booktitle={2021 IEEE International Conference on Robotics and Automation (ICRA)},
  pages={9877--9883},
  year={2021},
  organization={IEEE},
  DOI={10.1109/ICRA48506.2021.9561846}
}

@article{wang2024fin,
  title={Fin-bayes: A multi-objective bayesian optimization framework for soft robotic fingers},
  author={Wang, Xing and Wang, Bing and Pinskier, Joshua and Xie, Yue and Brett, James and Scalzo, Richard and Howard, David},
  journal={Soft Robotics},
  volume={11},
  number={5},
  pages={791--801},
  year={2024},
  publisher={Mary Ann Liebert, Inc.},
  DOI={https://doi.org/10.1089/soro.2023.0134}
}

@misc{jain2023multi,
  title={Multi-Objective GFlowNets}, 
      author={Moksh Jain and Sharath Chandra Raparthy and Alex Hernandez-Garcia and Jarrid Rector-Brooks and Yoshua Bengio and Santiago Miret and Emmanuel Bengio},
      year={2023},
      eprint={2210.12765},
      archivePrefix={arXiv},
      primaryClass={cs.LG},
      url={https://arxiv.org/abs/2210.12765}, 
}

@book{ehrgott2005multicriteria,
  title={Multicriteria optimization},
  author={Ehrgott, Matthias},
  volume={491},
  year={2005},
  publisher={Springer Science \& Business Media}
}

@article{deb2002fast,
  title={A fast and elitist multiobjective genetic algorithm: NSGA-II},
  author={Deb, Kalyanmoy and Pratap, Amrit and Agarwal, Sameer and Meyarivan, TAMT},
  journal={IEEE transactions on evolutionary computation},
  volume={6},
  number={2},
  pages={182--197},
  year={2002},
  publisher={Ieee}
}

@article{zhang2007moea,
  title={MOEA/D: A multiobjective evolutionary algorithm based on decomposition},
  author={Zhang, Qingfu and Li, Hui},
  journal={IEEE Transactions on evolutionary computation},
  volume={11},
  number={6},
  pages={712--731},
  year={2007},
  publisher={IEEE}
}

@article{konakovic2020diversity,
  title={Diversity-guided multi-objective bayesian optimization with batch evaluations},
  author={Konakovic Lukovic, Mina and Tian, Yunsheng and Matusik, Wojciech},
  journal={Advances in Neural Information Processing Systems},
  volume={33},
  pages={17708--17720},
  year={2020}
}

@article{brochu2010,
  title={A Tutorial on Bayesian Optimization of Expensive Cost Functions, with Application to Active User Modeling and Hierarchical Reinforcement Learning},
  author={Brochu, Eric Cora, Vlad. M and de Freitas, Nando},
  journal={arXiv},
  url= {https://doi.org/10.48550/arXiv.1012.2599},
  year={2010}
}

@book{rassmussen2006,
  title={Gaussian Processes for Machine Learning},
  author={Rasmussen, Carl Edward and Williams, Christopher K.I},
  publisher={MIT Press Direct},
  url= {https://doi.org/10.7551/mitpress/3206.001.0001},
  year={2005}
}

@article{rogersecfp2010,
    author = {Rogers, David and Hahn, Mathew},
    title = {Extended-Connectivity Fingerprints},
    journal = {Journal of Chemical Information and Modeling},
    volume = {50},
    number = {5},
    pages = {742-754},
    year = {2010},
    doi = {10.1021/ci100050t},
        note ={PMID: 20426451},
    URL = {https://doi.org/10.1021/ci100050t}
}

@inproceedings{gao2022pmo,
  author = {Gao, Wenhao and Fu, Tianfu and Sun, Jimeng and Coley, Connor W.},
  title = {Sample Efficiency Matters: A Benchmark for Practical Molecular Optimization},
  booktitle = {Advances in Neural Information Processing Systems (NeurIPS)},
  year = {2022},
  doi = {10.48550/arXiv.2206.12411}
}

@article{trippgregg2021,
    author = {Tripp, Austin and Simm, Gregor N. C. and  Hernandez-Lobato, Jose Miguel},
    title = {A Fresh Look at De Novo Molecular Design Benchmarks},
    journal = {NeurIPS Workshop 2021 AI4Science},
    year = {2021},
    URL = {https://openreview.net/pdf?id=gS3XMun4cl_}
}

@article{knowles2006parego,
  author = 	 {Joshua Knowles},
  title = 	 {ParEGO: A hybrid algorithm with on-line landscape approximation for expensive multiobjective optimization problems},
  journal = 	 {IEEE Transactions on Evolutionary Computation},
  year = 	 {2006},
  volume = 	 {10},
  number = 	 {1},
  pages = 	 {50-66},
}

@article{tchebysheff1976,   
    author = {Bowman Junior, V. J.}, 
    title = {On the Relationship of the Tchebycheff Norm and the Efficient Frontier of Multiple-Criteria Objectives}, 
    year = {1976}, 
    DOI = {10.1007/978-3-642-87563-2_5},
    journal = {Springer Science}, 
}

@inproceedings{smoothsettchebysheff2025,
  author = {Lin, Xi and Liu, Yilu and Zhang, XiaoYuan and Liu, Fei and Wang, Zhenkun and Zhang, Qingfu},
  title = {Few for Many: Tchebycheff Set Scalarization for Many-Objective Optimization},
  booktitle = {Proceedings of the International Conference on Learning Representations (ICLR)},
  year = {2025},
  doi = {10.48550/arXiv.2405.19650}
}

@inproceedings{emmerich2011hypervolume,
  author={Emmerich, Michael T. M. and Deutz, André H. and Klinkenberg, Jan Willem},
  booktitle={2011 IEEE Congress of Evolutionary Computation (CEC)}, 
  title={Hypervolume-based expected improvement: Monotonicity properties and exact computation}, 
  year={2011},
  pages={2147-2154},
  doi={10.1109/CEC.2011.5949880}}

@misc{daulton2020differentiable,
      title={Differentiable Expected Hypervolume Improvement for Parallel Multi-Objective Bayesian Optimization}, 
      author={Samuel Daulton and Maximilian Balandat and Eytan Bakshy},
      year={2020},
      eprint={2006.05078},
      archivePrefix={arXiv},
      primaryClass={stat.ML},
      url={https://arxiv.org/abs/2006.05078}, 
}

@article{hernandez-lobato-pesmo,
  title={Predictive Entropy Search for Multi-objective Bayesian Optimization},
  author={Hernandez-Lobato, Daniel. and Hernandez-Lobato, Jose Miguel and Shah Amar and Adams Ryan},
  journal={Proceedings of The 33rd International Conference on Machine Learning},
  volume={48},
  pages={1492-1501},
  year={2016}
}

@misc{tripp2024diagnosingfixingcommonproblems,
      title={Diagnosing and fixing common problems in Bayesian optimization for molecule design}, 
      author={Austin Tripp and José Miguel Hernández-Lobato},
      year={2024},
      eprint={2406.07709},
      archivePrefix={arXiv},
      primaryClass={cs.LG},
      url={https://arxiv.org/abs/2406.07709}, 
}

@misc{xie2023spaceexploredmeasuringchemical,
      title={How Much Space Has Been Explored? Measuring the Chemical Space Covered by Databases and Machine-Generated Molecules}, 
      author={Yutong Xie and Ziqiao Xu and Jiaqi Ma and Qiaozhu Mei},
      year={2023},
      eprint={2112.12542},
      archivePrefix={arXiv},
      primaryClass={cs.CE},
      url={https://arxiv.org/abs/2112.12542}, 
}

@article{hansenr2,
    author = {Hansen, Michael Pilegaard and Jaszkiewicz, Andrzej},
    year = {1998},
    month = {03},
    pages = {29},
    title = {Evaluating the quality of approximation to the non-dominated set},
    journal = {IMM Technical Report}
}

@inproceedings{fonsecahypervolume,
  author={Fonseca, C.M. and Paquete, L. and Lopez-Ibanez, M.},
  booktitle={2006 IEEE International Conference on Evolutionary Computation}, 
  title={An Improved Dimension-Sweep Algorithm for the Hypervolume Indicator}, 
  year={2006},
  volume={},
  number={},
  pages={1157-1163},
  doi={10.1109/CEC.2006.1688440}}

@article{brown2019guacamol,
   title={GuacaMol: Benchmarking Models for de Novo Molecular Design},
   volume={59},
   ISSN={1549-960X},
   url={http://dx.doi.org/10.1021/acs.jcim.8b00839},
   DOI={10.1021/acs.jcim.8b00839},
   number={3},
   journal={Journal of Chemical Information and Modeling},
   publisher={American Chemical Society (ACS)},
   author={Brown, Nathan and Fiscato, Marco and Segler, Marwin H.S. and Vaucher, Alain C.},
   year={2019},
   month=mar, pages={1096–1108} }

@article{lepage-mcintegration,
    title = {A new algorithm for adaptive multidimensional integration},
    journal = {Journal of Computational Physics},
    volume = {27},
    number = {2},
    pages = {192-203},
    year = {1978},
    issn = {0021-9991},
    doi = {https://doi.org/10.1016/0021-9991(78)90004-9},
    url = {https://www.sciencedirect.com/science/article/pii/0021999178900049},
    author = {G {Peter Lepage}},
}

@inproceedings{zitzler1998multiobjective,
    author="Zitzler, Eckart
    and Thiele, Lothar",
    editor="Eiben, Agoston E.
    and B{\"a}ck, Thomas
    and Schoenauer, Marc
    and Schwefel, Hans-Paul",
    title="Multiobjective optimization using evolutionary algorithms --- A comparative case study",
    booktitle="Parallel Problem Solving from Nature --- PPSN V",
    year="1998",
    publisher="Springer Berlin Heidelberg",
    address="Berlin, Heidelberg",
    pages="292--301",
}

@misc{golovin2020randomhypervolumescalarizationsprovable,
      title={Random Hypervolume Scalarizations for Provable Multi-Objective Black Box Optimization}, 
      author={Daniel Golovin and Qiuyi Zhang},
      year={2020},
      eprint={2006.04655},
      archivePrefix={arXiv},
      primaryClass={cs.LG},
      url={https://arxiv.org/abs/2006.04655}, 
}

@misc{paria2019flexibleframeworkmultiobjectivebayesian,
      title={A Flexible Framework for Multi-Objective Bayesian Optimization using Random Scalarizations}, 
      author={Biswajit Paria and Kirthevasan Kandasamy and Barnabás Póczos},
      year={2019},
      eprint={1805.12168},
      archivePrefix={arXiv},
      primaryClass={cs.LG},
      url={https://arxiv.org/abs/1805.12168}, 
}

@article{tanimoto_kernelsralaivola, 
  title = {Graph kernels for chemical informatics}, 
  author = {Ralaivola, Liva and Swamidass, S. J. and Saigo, H. and Baldi, P.},
  journal = {Neural Networks},
  year = {2005},
  volume = {18},
  number = {8},
  pages = {1093--1110},
  doi = {10.1016/j.neunet.2005.07.009},
}

@book{cohen2013statistical,
  title={Statistical power analysis for the behavioral sciences},
  author={Cohen, Jacob},
  year={2013},
  publisher={routledge}
}

@book{cliff2014ordinal,
  title={Ordinal methods for behavioral data analysis},
  author={Cliff, Norman},
  year={2014},
  publisher={Psychology Press}
}

\appendix

\section{Appendix}
\subsection{Final Performance Metrics after 200 BO evaluations}
\begin{table}[H]
\centering
\caption{Final hypervolume (mean $\pm$ std) after 200 BO evaluations for each MPO task, computed over 3 random seeds.}
\begin{tabular}{lcc}
\toprule
\textbf{Task} & \textbf{EHVI} & \textbf{Scalarized EI} \\
\midrule
Fexofenadine & $0.4022 \pm 0.0661$ & $0.3492 \pm 0.0190$ \\
Amlodipine   & $0.2421 \pm 0.0425$ & $0.2220 \pm 0.0251$ \\
Perindopril  & $0.2080 \pm 0.0016$ & $0.2088 \pm 0.0230$ \\
\bottomrule
\end{tabular}
\label{tab:final_hvi}
\end{table}

\begin{table}[H]
\centering
\caption{Final $R^2$ values (mean $\pm$ std) after 200 BO evaluations for each MPO task, computed over 3 random seeds.}
\label{tab:r2_values_std}
\begin{tabular}{lcc}
\toprule
\textbf{Task} & \textbf{EHVI} & \textbf{Scalarized EI} \\
\midrule
Fexofenadine & $0.3728 \pm 0.0204$ & $0.4360 \pm 0.0293$ \\
Amlodipine   & $0.1649 \pm 0.0203$ & $0.1816 \pm 0.0212$ \\
Perindopril  & $0.1582 \pm 0.0087$ & $0.1953 \pm 0.0322$ \\
\bottomrule
\end{tabular}
\end{table}

\subsection{Cohen's \texorpdfstring{$d$}{d} and Cliff's Delta}
\label{appendix2}
\begin{table}[H]
\centering
\caption{Effect size metrics comparing EHVI and scalarized EI on the hypervolume indicator across three MPO tasks.}
\begin{tabular}{lcc}
\toprule
\textbf{Task} & \textbf{Cohen's $d$} & \textbf{Cliff's Delta} \\
\midrule
Fexofenadine & 1.093 & 0.556 \\
Amlodipine   & 0.576 & 0.333 \\
Perindopril  & -0.050 & 0.333 \\
\bottomrule
\end{tabular}
\label{tab:hv_effect_sizes}
\end{table}

\begin{table}[ht]
\centering
\caption{Effect size between EHVI and scalarized EI for $R^2$ values across each task. Negative values indicate better performance by EHVI.}
\label{tab:r2_effectsize}
\begin{tabular}{lcc}
\toprule
\textbf{Task} & \textbf{Cohen's $d$} & \textbf{Cliff's Delta} \\
\midrule
Fexofenadine & $-2.560$ & $-1.000$ \\
Amlodipine   & $-0.770$ & $-0.556$ \\
Perindopril  & $-1.602$ & $-0.778$ \\
\bottomrule
\end{tabular}
\end{table}

\end{document}